%% file: acl2019.tex
\documentclass[11pt,a4paper]{article}
\usepackage[hyperref]{emnlp-ijcnlp-2019}
\usepackage{times}
\usepackage{latexsym}
\usepackage{graphicx}
\usepackage{amsfonts}
\usepackage{amsmath}
\usepackage{url}
\usepackage{tikz}
\usepackage{pgfplots}
\usepackage{verbatim}
\usepackage{multirow}
\usepackage{booktabs}
\usepackage{subfigure}
\usepackage{algorithm}
\usepackage{algorithmicx}
\usepackage{algpseudocode}

\aclfinalcopy 

\title{Collaborative Policy Learning for Open Knowledge Graph Reasoning}

\author{Cong Fu$^{1*}$, Tong Chen$^2$\thanks{~~~Work done while the authors interned at USC.}, Meng Qu$^3$, Woojeong Jin$^4$, Xiang Ren$^4$ \\
  $^1$Zhejiang University, $^2$Carnegie Mellon University, $^3$University of Montreal \\$^4$University of Southern California \\
  {fc731097343@gmail.com}, tongc2@andrew.cmu.edu, meng.qu@umontreal.ca\\
  \{woojeong.jin, xiangren\}@usc.edu \\}

\begin{document}
\maketitle

\begin{abstract}
  \input{0-abs.tex}

\end{abstract}

\input{1-intro.tex}
\input{2-framework.tex}

\input{3-models.tex}
\input{4-exp.tex}

\input{5-related.tex}
\input{6-con.tex}

\bibliography{acl2019}
\bibliographystyle{acl_natbib}
\clearpage
\appendix

\input{Supplemental.tex}

\end{document}

%% file: 0-abs.tex

In recent years, there has been a surge of interests in interpretable graph reasoning methods. However, these models often suffer from limited performance when working on sparse and incomplete graphs, due to the lack of evidential paths that can reach target entities. Here we study \textit{open knowledge graph reasoning}---a task that aims to reason for missing facts over a graph augmented by a background text corpus.
A key challenge of the task is to filter out ``\textit{irrelevant}" facts extracted from corpus, in order to maintain an effective search space during path inference. We propose a novel reinforcement learning framework to train two collaborative agents jointly, i.e., a \textit{multi-hop graph reasoner} and a \textit{fact extractor}. The fact extraction agent generates fact triples from corpora to enrich the graph on the fly; while the reasoning agent provides feedback to the fact extractor and guides it towards promoting facts that are helpful for the interpretable reasoning. Experiments on two public datasets demonstrate the effectiveness of the proposed approach. Source code and datasets used in this paper can be downloaded at \url{https://github.com/shanzhenren/CPL}.

%% file: 1-intro.tex

\section{Introduction}
\begin{figure}[tb]
\centering
\vspace{-0.1cm}
\includegraphics[width=0.84\columnwidth]{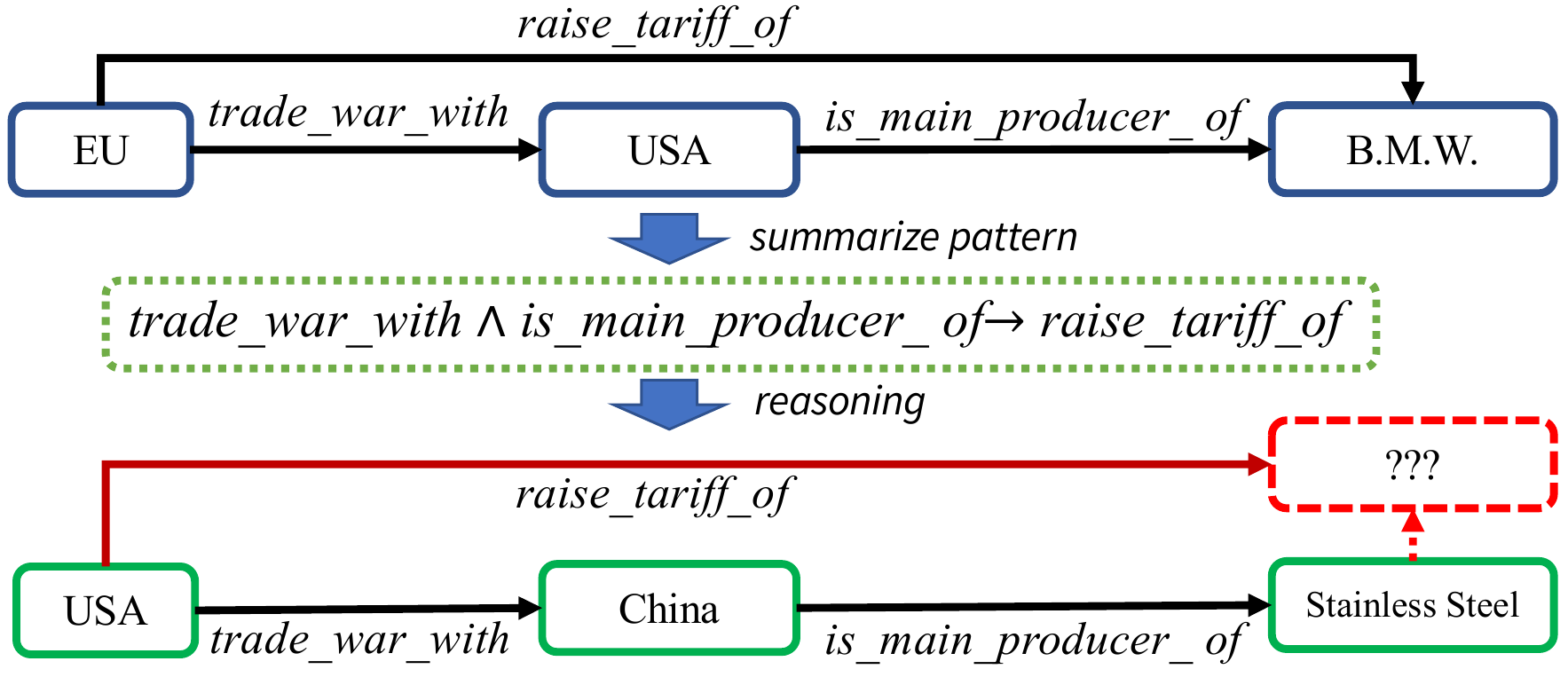}
\vspace{-0.2cm}
\caption{\textbf{Illustration of the Knowledge Graph Reasoning Task.} Given an entity (e.g., \textit{Miami}) and a query relation (e.g., \textit{located in}), we learn to infer reasoning paths over the existing graph structure to help predict the answer entity (i.e., \textit{USA}).}
\label{fig:GRexample}
\vspace{-0.2cm}
\end{figure}

Knowledge graph completion or reasoning---i.e., the task of inferring the missing facts (entity relationships) for a given graph---is an important problem in natural language processing and has a wide range of applications~\cite{TransE:11, socher2013reasoning, ComplEx:16}.
Recent neural graph reasoning methods, such as MINERVA~\cite{GFAW:17}, DeepPath \cite{DeepPath:17} and Multi-Hop~\cite{multi-lin2018}, have achieved impressive results on the task, offering both good \textit{prediction accuracy} (compared to embedding-based methods~\cite{ComplEx:16, conve:18}) and \textit{interpretability} of the model predictions. These reasoning methods frame the link inference task as a path finding problem over the graph (see Fig.~\ref{fig:GRexample} for example).

\begin{figure*}[ht]
\centering
\vspace{-0.1cm}
\includegraphics[width=0.9\linewidth]{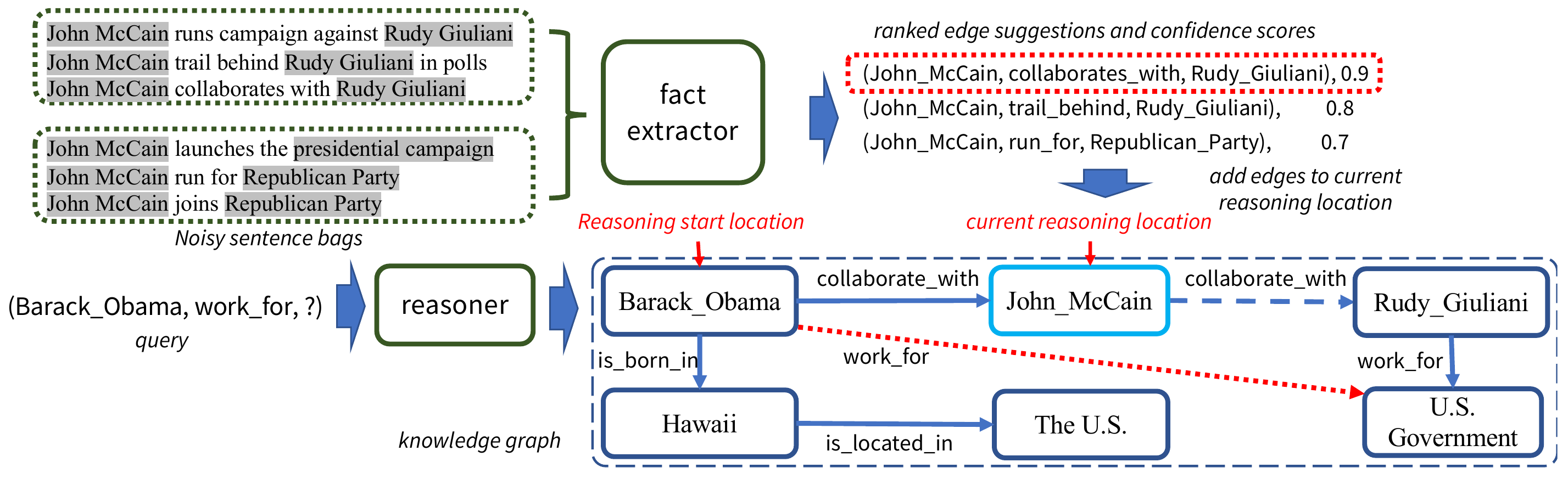}
\vspace{-0.1cm}
\caption{\textbf{Overview of our CPL framework for the OKGR problem.} To augment the reasoning with the information from a background corpus, CPL extracts relevant facts (e.g., the triple linked by the blue dotted arrow) to augment the KG dynamically. CPL involves two agents: one learns fact extraction policy to suggest relevant facts; the other learns to reason on dynamically augmented graphs to make predictions (e.g., the red dotted arrow).}
\label{fig:Motivation}
\vspace{-0.2cm}
\end{figure*}
However, current neural graph reasoning methods encounter two main challenges as follows: 
(1) their performance are often sensitive to the sparsity and completeness of the graph---missing edges (i.e., potential false positives) make it harder to find evidential paths reaching target entities.
(2) existing models assume the graph is static, and cannot adapt to dynamically enriched graphs where emerging new facts are constantly added.

In this paper, we study the new task of Open Knowledge Graph Reasoning (OKGR), where the new facts extracted from the text corpora will be used to augment the graph dynamically while performing reasoning (as illustrated in Figure \ref{fig:Motivation}). All the recent joint graph and text embedding methods focus on learning better knowledge graph embeddings for reasoning \cite{rcnet, OpenNRE:18}, but we consider adding more facts to the graph from the text to improve the reasoning performance and further provide interpretability. 
A straightforward solution for the OKGR problem is to directly add extracted facts (by a pre-trained relation extraction model) to the graph. However, most facts so extracted may be noisy or irrelevant to the path inference process. Moreover, adding a large number of edges to the graph will create an ineffective search space and cause scalability issues to the path finding models.
Therefore, it is desirable to design a method that can filter out irrelevant facts for augmenting the reasoning model.
To address the above challenges for OKGR, we propose a novel \textit{collaborative policy learning} (CPL) framework to jointly train two RL agents in a mutually enhancing manner. In CPL, besides training a \textit{reasoning} agent for path finding, we further introduce a \textit{fact extraction} agent, which learns the policy to select relevant facts extracted from the corpus, based on the context of the reasoning process and the corpus (see Fig.~\ref{fig:Motivation}). At inference time, the fact extraction agent dynamically augments the graph with only the most informative edges, and thus enables the reasoning agent to identify positive paths effectively and efficiently.
 
Specifically, during policy learning, the reasoning agent will be rewarded when reaching the targets, while this positive feedback will also be transferred back to the fact extracting agent if its edge suggestions are adopted by the reasoning agent, i.e. making up correct reasoning paths. This ensures that the fact extraction policy can be learned in a way that edges which are beneficial to path inference will be preferred. By doing so, the fact extraction agent can learn to augment knowledge graphs dynamically to facilitate the reasoning agent, while the reasoning agent performs effective path-inference and provides reward signals to the fact extraction agent. Please refer to Sec. \ref{prop_frame} and Fig. \ref{fig:JointModel} for more implementation details.

The major contributions of our work are as follows: (1) We study knowledge graph reasoning in an ``open-world" setting, where new facts extracted from background corpora can be used to facilitate path finding; 
(2) We propose a novel collaborative policy learning framework which models the interactions between fact extraction and graph reasoning; 
(3) Extensive experiments and analysis are conducted to demonstrate the effectiveness and strengths of our proposed method.


%% file: 2-framework.tex
\section{Background and Problem}
\label{background}
This section introduces basic concepts and notations related to the knowledge graph reasoning task and provides a formal problem definition.

A \textit{knowledge graph} (KG) can be represented by a set of triples (facts) $G=\{(e_s,r,e_o)|e_s,e_o \in E, r \in R\}$, where $E$ is the set of entities and $R$ is the set of relations. $e_s$, $r$, and $e_o$ are the subject entity, relation, and object entity respectively. 

The task of \textit{knowledge graph reasoning} (KGR) is defined as follows. Given the KG $G$, a query triple ($e_s, r_q, e_q$) where $e_q$ is unknown, KGR is to infer $e_q$ through finding a path starting from $e_s$ to $e_q$ on $G$, namely $\{(e_s, r_1, e_1), ... , (e_n, r_n, e_q)\}$. Usually, KGR methods produce multiple candidate answers by ranking the found paths, while traditional KG completion methods rank all possible answer triples by exhaustively enumerating. 


A \textit{background corpus} is a set of sentences labeled with respective entity pairs, namely $C=\{(s_i :(e_k, e_j))|s_i \in S, e_k,e_j \in E\}$, where $S$ is the set of sentences, and the corpus shares the same entity set with $G$. In our problem setting, we assume the entities have already been extracted; thus, extracting facts from the corpus is equivalent to the relation extraction task. We process the corpus by labeling the sentences with subject and object entity pairs through Distant Supervision \cite{Mintz:09}. There may be many sentences labeled with the same entity pair. Following the formulation of previous work \cite{PCNN:16}, we organize the sentences into \textit{sentence bags}, i.e., A sentence bag contains the sentences which are labeled with the same entity pair.

\smallskip
\noindent
\textbf{Problem.}
Formally, \textit{Open Knowledge Graph Reasoning} (\textbf{OKGR}) aims to perform KGR based on both $G$ and $C$, where $G$ is dynamically enriched by the facts extracted from $C$. This paper focuses on OKGR, i.e., empowering KGR with the corpus information and enriching the graph with relevant facts dynamically. Thus, the evaluation of the relation extraction performance are out of the scope of this paper. We leave this as future work.

\section{Proposed Framework}
\label{prop_frame}
\vspace{-0.2cm}
\noindent
\textbf{Overview.}
To resolve the challenges in OKGR, we propose a novel collaborative policy learning (CPL) framework (see Fig.~\ref{fig:Motivation}), which jointly train two RL agents, i.e., a path reasoning agent and a fact extraction agent.
Given a query ($e_s, r_q, e_q$), the reasoning agent tries to infer $e_q$ via finding a reasoning path on the (augmented) $G$, while the fact extraction agent aims to select the most informative facts from $C$ to enrich $G$ dynamically. With such an extractor, the framework can effectively overcome the edge sparsity problem while remaining reasonably efficient (compared to the naive solution that adds all possible facts to $G$). We train the extraction agent by rewarding it according to the the reasoning agent's performance. Hence the fact extractor can learn how to extract the most informative facts to benefit the reasoning.

\subsection{Graph Reasoning Agent}
\begin{figure*}[ht]
\centering
\includegraphics[width=0.84\linewidth]{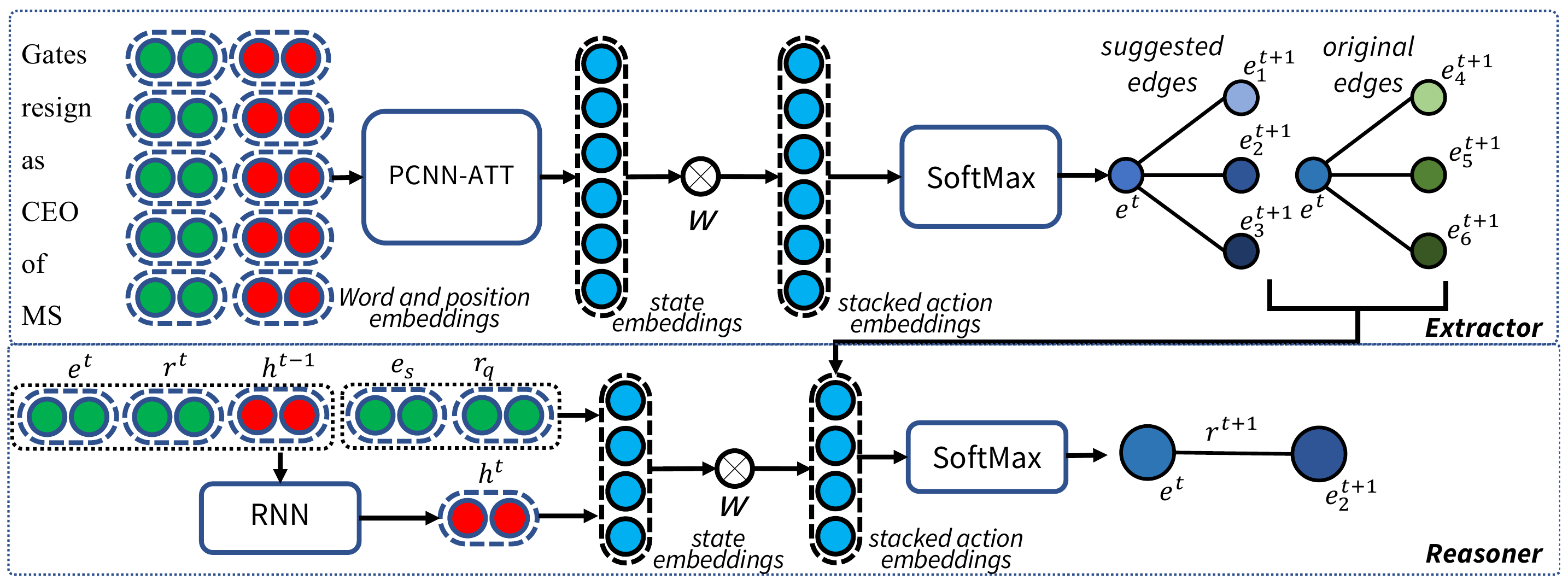}
\vspace{-0.1cm}
\caption{\textbf{Detailed Model Design of Collaborative Policy Learning (CPL).} Take PCNN-ATT as an example of the sentence encoder. The figure shows how it works at a certain inference time step $t$, the reasoner is at entity $e^t$ and will select one edge from the \textit{joint action space}, which consists of new edges extracted by the extractor and the edges in the original graph.}
\vspace{-0.2cm}
\label{fig:JointModel}
\end{figure*}

The goal of the reasoner is learning to reason via paths finding on KGs. Specifically, given $e_s, r_q$, the reasoner aims at inferring a path from $e_s$ to some entity $e_o$ regarding $r_q$, and specifying how likely the relationship $r_q$ holds between $e_s$ and $e_o$. The inference path acts as the evidence of the prediction, and thus offers interpretation (see Fig. \ref{fig:GRexample}). At each time step, the reasoner tries to select an edge based on the observed information. The Markov Decision Process (MDP) of the reasoner is defined as follows:

\smallskip
\noindent
\textbf{State.} In path-based reasoning, each succeeding edge is closely related to the preceding edge on the path and the query in semantics. Similar to MINERVA\cite{GFAW:17}, we want the state to encode all observed information, i.e., we define $s^t_R = (e_s, r_q, h^t) \in \mathcal{S}_R$, where $h^t$ encodes the path history, and ($e_s$, $r_q$) is the context shared among all states. Specifically, we use a LSTM module to encode the history, $h^t =$ LSTM$(h^{t-1}, [r^t, e^t])$ (see Fig. \ref{fig:JointModel}). $e^t$ is the current reasoning location and $r^t$ is the previous relation connecting $e^t$.

\smallskip
\noindent
\textbf{Action.} At time $t$, the reasoner will select an edge among $e^t$'s out-edges. The reasoner's action space is a union of the edges in the current KG and edges extracted from the corpus. See Sec. \ref{collabor-detail} for details.

\smallskip
\noindent
\textbf{Transition.} The transition function $f:\mathcal{S}_R\times \mathcal{A}_R\rightarrow \mathcal{S}_R$ is defined as $f(s^t_R, a^t_R) = (e^{s}, r^{q}, h^{t+1})$ naturally.

\smallskip
\noindent
\textbf{Reward.} The reasoner is expected to learn effective reasoning path patterns. We let it explore for a fixed number of steps, which is a hyper-parameter. Only when it reaches the correct target entity, it receives a terminal reward 1, and 0 otherwise. All intermediate states always receive reward 0.

\vspace{-0.1cm}

\subsection{Fact Extraction Agent}
The fact extractor learns to suggest the most relevant facts w.r.t. the current inference step of the reasoner. Suppose the reasoner arrives at entity $e^t$ on the graph at time $t$, the fact extractor will extract facts in the form of $(e^t, r', e') \notin G$ from the corpus and add them to the graph temporarily. Consequently, the reasoner is offered more choices to expand the reasoning path. 

\smallskip
\noindent
\textbf{State.} When the reasoner is at $e^t$, the fact extractor tries to extract information from the corpus (sentences) and suggests promising out-edges of entity $e^t$. Let $b_{e^t}$ denote the sentence bags labeled with $(e^t, e'), e'\in E$. We define the state of the fact extractor to encode the current observed information, i.e., $s^t_E = (b_{e^t}, e^t) \in \mathcal{S}_E$, where $\mathcal{S}_E$ is the whole state space, containing all possible combinations of entities and corresponding sentence bags.

\smallskip
\noindent
\textbf{Action.} The goal of the fact extractor is to select a reasoning-relevant fact contained in the corpus semantically. At step $t$, the reasoner will move to a new entity from $e^t$, and hence only the out-edges of $e^t$ should be considered, e.g., $(e^t, r', e')$ (see Fig. \ref{fig:JointModel}). Therefore, for each fact $(e^t, r', e')$ which can be extracted from the sentence bag $b_{e^t}$, we can derive an action $a^t_E = (r', e')$, and the action space at step $t$ can be denoted as $A^t_E =\{(r',e')\}_{(e^t, r', e') \in b_{e^t}} \subset \mathcal{A}_E$. $\mathcal{A}_E$ is the whole action space containing all facts in the corpus.

\smallskip
\noindent
\textbf{Transition.} The transition function $f:\mathcal{S}_E\times \mathcal{A}_E\rightarrow \mathcal{S}_E$ is defined as $f(s^t_E, a^t_E) = (r', e')$. 

\smallskip
\noindent
\textbf{Reward.} The fact extractor receives a step-wise delayed reward from the reasoner according to how it improves the reasoners performance. The extractor will be positively rewarded when its suggestion benefits the reasoning process. Please see Sec. \ref{collabor-detail} for details.

\subsection{Collaborative Policy Learning}
\label{collabor-detail}

In this section, we will introduce the detailed training process and the collaboration mechanism between the two agents.
At the high level, we adopt an \textit{alternative training procedure} to update the two agents jointly: training one of the agents for a few iterations while freezing the other; and vice versa. The policies of both agents are updated via REINFORCE algorithm~\cite{Reinforce:1999} (details are in later of this section). Specifically, we introduce the details on agent collaboration as follows.

\smallskip
\noindent
\textbf{Augmented Action Space for Reasoning.} 
At time $t$, the fact extractor helps the reasoner via expanding its action space with new edges extracted from the corpus (see Fig. \ref{fig:JointModel}). Due to the sparsity and incompleteness of the KG, there may be missing edges preventing the reasoner from inferring the correct reasoning path (Fig. \ref{fig:Motivation}). Therefore, we add high-confidence edges extracted by the extractor to the action space of the reasoner. Formally, at time $t$, the reasoner is at location $e^t$ (Fig. \ref{fig:JointModel}) and tries to select an edge out of all out-edges of $e^t$. Let $A_K^t$ denote the edge set in the current KG, $A_K^t = \{(r,e)|(e^t, r, e) \in G\}$. Let $A_C^t$ denote the edge set suggested by the extractor, $A_C^t = \{(r',e')|(e^t, r', e') \in C\}$. The action space at time $t$ of the reasoner is defined as $A^t_R = A_K^t \cup A_C^t$, $A^t_R \subset \mathcal{A}_R$, where $\mathcal{A}_R$ denotes the whole action space of the reasoner, i.e., all possible edges in the KG and the corpus. The reasoner learns a policy to select the best edges out of the joint action space for reasoning.

\smallskip
\noindent
\textbf{Reasoning Feedback for Fact Extraction.} 
The reasoner helps the extractor to learn the extracting policy through providing feedbacks regarding how much the extractor contributes to the reasoning. Therefore we define that the fact extractor receives a step-wise delayed reward from the reasoner. Specifically, when the reasoner finishes exploration (at time $T$) and arrives at the correct target, we consider the path is effective and positive for reasoning. If the fact extractor contributes to this positive path, it can be rewarded positively, i.e., if an edge on the positive path is suggested by the extractor at time $t, 0\le t\le T$, the extractor will be rewarded 1 at time $t$, and 0 otherwise. Extracted edges triggering positive rewards will be kept in the graph, while the others will be removed when both agents move to the next state.

\smallskip
\noindent
\textbf{Policy Update.} The MDPs of both agents are explicitly defined above now, and we can use the typical REINFORCE algorithm \cite{Reinforce:1999} to train the two agents. Specifically, their goals are maximizing the reward expectation, defined as 
\begin{equation}
\begin{small}
    J(\theta) = \mathbb{E}_{\pi_{\theta}(a|s)}[R(s,a)],
\end{small}
\label{equ:reward}
\end{equation}
where $R(s,a)$ is the reward of selecting $a$ given $s$, and $\pi_{\theta}(a|s)$ is the policy learned by the agents and will be defined formally in Section. \ref{implementation}.

Given a training sequence sampled from $\pi_{\theta}$: $\{(s^1,a^1,r^1), ... , (s^T, a^T, r^T)\}, r^t = R(s^t,a^t)$ , at time step $t$, the parameters are updated according to the REINFORCE algorithm: 
\vspace{-0.1cm}
\begin{equation}
\scalebox{0.8}{$
\begin{split}
    \theta \leftarrow &\theta + \alpha\nabla_{\theta}\log\pi_{\theta}(a^t|s^t)G^t\\
    G^t &= \sum_{k=t}^{T}\gamma^{k-t}R(s^{k},a^{k}),
\end{split}
$}
\label{equ:update}
\end{equation}
where $G^t$ is the discounted accumulated reward. 

According to Eq.~(\ref{equ:update}), we can see that REINFORCE will update the parameters only when $G^t$ is non-zero. In other words, the value of $\gamma$ determines how the parameters are updated and to what extent the the internal states will be influenced by the future. If $\gamma>0$, for positive training sequences, it is easy to verify that the $G^t$ of all states will be non-zero. Thus, the internal states will be positively rewarded, and the model parameters will be updated by the gradients of the internal states. For different task, we should carefully select the value of $\gamma$. For the extractor, we set $\gamma=0$ for the extractor to avoid policy updating on zero-rewarded state-action experiences. This is because zero-rewarded experiences are mostly negative examples. Specifically, if a state of the extractor is zero-rewarded, we can infer that either the suggested edge is not selected by the reasoner, or the selected edge does not contribute to reaching the target. We can not allow the model to be updated on such experiences, so we set $\gamma=0$ to avoid the influence of future. In contrast, we set $\gamma=1$ for the reasoner because all the intermediate selected edges are meaningful as long as it leads to the target finally.


%% file: 3-models.tex
\section{Model Implementation}
\label{implementation}
In this section, we introduce the policy network architectures (cf. Fig.~\ref{fig:JointModel}) of the two agents and provide details on model training and inference.

\subsection{Policy Network Architectures}
\noindent
\textbf{Reasoning Agent.} We construct the state embedding by concatenating all related embeddings, $s^t_R = [e_s,r_q,h^t]$, where $h^t =$ LSTM $(h^{t-1},[r^t, e^t])$. We construct the action embedding by concatenating the relation-entity embedding pair, i.e., $a^t_R = [r,e], (e^t, r,e) \in G$ $\cup$ $C$. We stack all action embeddings in $\mathbf{A}^t_R$. The policy network is defined as:
\begin{equation*}
\begin{small}
    \pi_{\theta}(a^t_R|s^t_R) = \sigma(\mathbf{A}^t_RW_2(Relu(W_1[e_s,r_q,h^t]))),
\end{small}
\end{equation*}
$\sigma$ is softmax, and $W_1$, $W_2$ are learnable weights.

\smallskip
\noindent
\textbf{Fact extraction Agent.}
We use a PCNN-ATT as the sentence encoder in our experiments to construct the distributed representations for the sentences. Let $b_{e^t}$ denote all the sentence bags labeled by $(e^t, e'), e' \in E$. At time $t$, we input $b_{e^t}$ into the PCNN-ATT to obtain the sentence-bag-level embeddings $E_b^t$, which is regarded as the latent state embeddings. As mentioned in Sec. \ref{background}, the object entity has been labeled beforehand. We need to select the best relation first, then select the best entity under this relation. Thus, we stack the relation embeddings in $A^t_E$ as $\mathbf{A}^t_E\in \mathbb{R}^{|A^t_E|\times d}$, where $d$ is the dimension of relation embedding. The policy network is defined formally as:
\begin{equation*}
\begin{small}
    \pi_{\theta}(a^t_E|s^t_E) = \sigma(\mathbf{A}_E^tWE_b^t),
\end{small}
\end{equation*}
where $W$ is a learnable weight. 

The extractor will predict the scores for each sentence bag regarding the relation. We will select the sentence bag with the highest score, more formally, the corresponding relation-entity pair in that sentence bag will be chosen as the next action. We train the agents as introduced in Sec. \ref{collabor-detail}.


\subsection{Model Training and Inference}
\noindent
\textbf{Training.}
We use \textit{model pre-training} and \textit{adaptive sampling} to increase training efficiency. In particular, we first train the reasoner on the original KG to get a better initialization. Similarly, we train the extractor on the corpus labeled by distant supervision. Next, we use adaptive sampling to adaptively increase the selecting-priority of corpus-extracted-edges when generating training experiences for the two agents. Adaptive sampling is designed to encourage the reasoner to explore more on new edges and facilitate the collaboration during the joint training. Replay memories~\cite{replaymem} are also used to increase training efficiency. We develop several model variants such as removing adaptive sampling or replay memory, or freezing the extractor all the time to conduct ablation studies. Please see Supplementary Material for more details. 

  \begin{table*}[tb]
    \centering
    \small
    \scalebox{0.88}{
    \begin{tabular}{ccccccccccc}
         \hline
         Dataset & \#triples(C) & \#triple(G) & \#entities(C) & \#entities(G) & \#rel(C) & \#rel(G) &S(train) &S(test) & CT/CE & CR/KR\\
         \hline
         FB60K-NYT10  & 172,448 & 268,280 &63,696 & 69,514 &57&1,327&570k & 172k & 2.71 & 0.04\\
         UMLS-PubMed  & 910,320 & 2,030,841 &6,575&59,226 &271&443   & 4,73M & 910k&138.45&0.61 \\
         \hline
    \end{tabular}}
    \vspace{-0.1cm}
    \caption{\textbf{The dataset information.} \#triples(C) \& \#triples(G) denote the number of triples in the corpus and the KG respectively, and so on. S(train) denotes the number of sentences in the training corpus, while S(test) denotes the number of sentences in the testing corpus. CT/CE denotes triple-entity ratio. Lower triple-entity ratio indicates less triples per entity in average can be extracted from the corpus. CR/KR denotes corpus-relation-quantity/KG-relation-quantity ratio. Lower CR/KR indicates less information overlap between the corpus and the KG.}
    \label{tab:dataoverlap}
\end{table*}

\smallskip
\noindent
\textbf{Inference.}
At inference (reasoning) time, we use the trained model to predict missing facts via path finding. The process is similar to the training experience generation step in the training stage, i.e. using the reasoner for path-inference while the extractor suggests edges from the corpus constantly. The only differences are that we do not request rewards, and we use beam search to generate multiple reasoning paths over the graph, and rank them by the scores from the reasoner.


%% file: 4-exp.tex
\section{Experiment}


\subsection{Datasets and Compared Methods}
\label{data-alg}

\begin{table}[ht!]
    \centering
    \vspace{-0.4cm}
    \scalebox{0.7}{
    \begin{tabular}{cc}
    \toprule
    Dataset & Relations   \\
    \midrule
    \multirow{8}{*}{UP} 
    & \textit{'gene\_associated\_with\_disease'}\\
    & \textit{'disease\_has\_associated\_gene'}\\
    & \textit{'gene\_mapped\_to\_disease'}\\
    & \textit{'disease\_mapped\_to\_gene'}\\
    & \textit{'may\_be\_treated\_by'}\\
    & \textit{'may\_treat'}\\
    & \textit{'may\_be\_prevented\_by'}\\
    & \textit{'may\_prevent'}\\
    \midrule
    \multirow{16}{*}{FN}
    & \textit{'people/person/nationality'}\\  
    & \textit{'location/location/contains'}\\
    & \textit{'people/person/place\_lived'}\\
    & \textit{'people/person/place\_of\_birth'}\\
    & \textit{'people/deceased\_person/place\_of\_death'}\\
    & \textit{'people/person/ethnicity'}\\
    & \textit{'people/ethnicity/people'}\\
    & \textit{'business/person/company'}\\
    & \textit{'people/person/religion'}\\
    & \textit{'location/neighborhood/neighborhood\_of'}\\
    & \textit{'business/company/founders'}\\
    & \textit{'people/person/children'}\\
    & \textit{'location/administrative\_division/country'}\\
    & \textit{'location/country/administrative\_divisions'}\\
    & \textit{'business/company/place\_founded'}\\
    & \textit{'location/us\_county/county\_seat'}\\
    
    \bottomrule
    \end{tabular}
    }
    \caption{\textbf{The concerned relations in two datasets.} UP means the UMLS-PubMed dataset, while FB means the FB60K-NYT10 dataset. 
    }
    \vspace{-0.4cm}
    \label{tab:relations}
\end{table}

\noindent
\textbf{Datasets.} 
We construct two datasets for evaluation: \textit{FB60K-NYT10}\footnote{\small\url{https://github.com/thunlp/OpenNRE}} and \textit{UMLS-PubMed}\footnote{\small\url{http://umlsks.nlm.nih.gov/}\\\url{https://www.ncbi.nlm.nih.gov/pubmed/}}. FB60K-NYT10 dataset includes the FB-60K KG and the NYT10 corpus; The UMLS dataset contains the UMLS KG and the PubMed corpus. Statistics of both datasets are summarized in Table~\ref{tab:dataoverlap}. We study the datasets and find that the relation distributions of the two datasets are very imbalanced. There are not enough reasoning paths for some relation types. Moreover, some relations are meaningless and of no reasoning value. Thus, we select a few meaningful and valuable relations (suggested by domain experts, in Table \ref{tab:relations}) with enough reasoning paths and construct two sub-graphs accordingly. To show the impact of graph size, we sub-sample the KG into different sub-graph. Specifically, for the two datasets, we first partition the whole KG into three parts according to the proportion 8:1:1 (The training set, validation set, and testing set). Next we create sub-train-sets with different ratio via random sampling.

\smallskip
\noindent
\textbf{Analysis of corpus-KG alignment.}
We analyze the information overlap (i.e., alignment) between the corpus and the KG in Table \ref{tab:dataoverlap}. The CT/CE (the ratio of triple quantity against entity quantity) of PubMed is far higher than NYT10. Higher CT/CE indicates adding corpus-edges to the KG increases the average degree more significantly, leading to more reduction in sparsity. The low CR/KR ratio of FB60K-NYT10 indicates the overlap between FB60K and NYT10 is lower than that between UMLS and PubMed. We can conclude that the alignment level of FB60K-NYT10 is lower than UMLS-PubMed. Intuitively, FB60K-NYT10 is a more difficult dataset than UMLS-PubMed.

\begin{table*}[ht]
\centering
\scalebox{0.78}{
\begin{tabular}{lcccccccc}
\hline
Model / Dataset    & \multicolumn{2}{c}{\textbf{20\%}} & \multicolumn{2}{c}{\textbf{40\%} } & \multicolumn{2}{c}{\textbf{70\%}} & \multicolumn{2}{c}{\textbf{100\%}} \\ 
          & Hits@5        & Hits@10        & Hits@5        & Hits@10        & Hits@5         & Hits@10   & Hits@5         & Hits@10     \\ \hline\hline
TransE~\cite{TransE:11}&7.12& 11.17&26.86 & 38.08& 31.32&43.58 &32.28 &45.52 \\
DisMult~\cite{DistMult:15}&14.66&21.16&26.90 &38.35 &31.65 &44.98 &32.80 & 47.50 \\
ComplEx~\cite{ComplEx:16}&18.58& 18.18& 23.77&34.15 &30.04 &43.60 &31.84 & 46.57\\
ConvE~\cite{conve:18}& {20.51} & {30.11}& \textbf{28.01} &\textbf{42.04}& 31.01 &45.81&30.35&45.35 \\
RotatE~\cite{sun2019rotate} &4.03 &6.50 &8.65 &13.21 &14.90 &21.67 &20.75 &27.82 \\
\hline
RC-Net~\cite{rcnet}&7.94&10.77&7.56&11.43&8.31 & 11.81&9.26&12.00 \\ 
TransE+LINE~&\textbf{23.63}&\textbf{31.85}&24.86&38.58 &25.43 &34.88 &22.31&33.65 \\
JointNRE~\cite{OpenNRE:18}&21.05 &31.37 &27.96 &40.10 &30.87 &44.47 &- &- \\ 
\hline
MINERVA~\cite{GFAW:17}&11.55&19.87&24.65 &35.71 &35.8 &46.26 & 57.63&63.83\\
Two-Step &8.37 &13.5 &22.75 &32.79 &33.14 & 43.35 &55.59&63.49\\
CPL (our method) & 15.32 & 24.22 & 26.96 & 38.03 & \textbf{37.23} & \textbf{47.60} & \textbf{58.10}&\textbf{65.16}\\ \hline
\end{tabular}
}
\vspace{-0.1cm}
\caption{\textbf{Performance comparison on the KG reasoning on the UMLS-PubMed dataset.\footnote{We cannot obtain results for Multi-Hop~\cite{multi-lin2018} as the published code gave out-of-memory error (due to large entity space). JointNRE also fails on the 100\% partition due to out-of-memory.}} We test on different graph sizes (i.e., 20-100\% of the original graph) using Hits@K (in \%). CPL is the best performing graph reasoning method, and gradually outperforms the others when the graphs are denser.}
\vspace{-0.1cm}
\label{overall-umls}
\end{table*}
\smallskip
\noindent
\textbf{Compared Algorithms.} 
We compare our algorithm with (1) SOTA methods for KG embedding ; (2) methods for joint text and graph embedding; and (3) neural graph reasoning methods.

For triple-ranking-based KG embedding methods, we evaluate \textbf{DistMult} \cite{DistMult:15}, \textbf{ComplEx} \cite{ComplEx:16}, and \textbf{ConvE} \cite{conve:18}. 
For joint text and graph embedding methods, we evaluate \textbf{RC-Net}~\cite{rcnet} and \textbf{Joint-NRE}~\cite{OpenNRE:18}. We also construct a baseline,
\textbf{TransE+LINE}, by constructing a word-entity co-occurrence network as RC-Net does. We use LINE~\cite{tang2015line} and TransE \cite{TransE:11} to jointly learn the entity and relation embeddings to preserve the structure information within the co-occurrence network and the KG.
For neural graph reasoning method, we use
\textbf{MINERVA}~\cite{GFAW:17}, a reinforcement learning based path reasoning method\footnote{\small There are other SOTA path-based knowledge reasoning methods such as Multi-Hop \cite{multi-lin2018} and DeepPath \cite{DeepPath:17} However, DeepPath needs extra path-level supervisions which we do not have and Multi-Hop suffers from out-of-memory problems on large-scale datasets.}. 

To validate the effectiveness of fact extraction policy in CPL, we design a two-step baseline (i.e., \textbf{Two-Step}).  It first uses PCNN-ATT to extract relational triples from the corpora, and augments KG with the triples whose prediction confidences are greater than a threshold. PCNN-ATT \cite{PCNN:16} is a fact extraction model, which completes the fact extraction part. We tune the threshold on the dev-set. Then, a MINERVA model is trained on the augmented KG for reasoning.

\textbf{CPL} is our full-fledged model as introduced in Sec.~\ref{prop_frame}. For all the methods, we upload the source codes and list hyper-parameters we used in the supplemental materials.

\subsection{Evaluation and Experimental Setup}
Following previous work on KG completion~\cite{TransE:11}, 
we use Hits@K and mean reciprocal rank (MRR) to evaluate the effectiveness of the KGR and OKGR.
Given a query $(e_s, r, ?)$ (for each triple in the test set, we pretend not to know the object entity), we rank the correct entity $e_q$ among a list of candidates entities. Suppose $rank_i$ is the rank of the correct answer entity for the $i^{th}$ query. We define \begin{small}$Hit@K = \sum_i{\mathbf{1}(rank_i<K)}/N$\end{small}
and \begin{small}$MRR = \frac{1}{N}\sum_i\frac{1}{rank_i}$\end{small}.

In our experiments, we use a held-out validation set for all compared methods to search for the best hyper-parameters and the best model for testing (via logging checkpoints).
For all methods, we train models using three fixed random seeds (55, 83, 5583), and report the metrics in average. More details on model training can be found in the Supplementary Material.

\begin{table*}[tb]
\centering
\scalebox{0.78}{
\begin{tabular}{lcccccccccccc}
\hline
Model / Dataset     & \multicolumn{3}{c}{\textbf{20\%}} & \multicolumn{3}{c}{\textbf{50\%}} & \multicolumn{3}{c}{\textbf{100\%}} \\ 
   &    Hits@5        & Hits@10  &MRR  &    Hits@5        & Hits@10  &MRR   &     Hits@5        & Hits@10  &MRR      \\ 
\hline\hline
TransE~\cite{TransE:11} & 15.12 & 18.83  & 12.57 & 19.38& 23.2& 13.36  &38.53&43.38&29.90    \\
DisMult~\cite{DistMult:15} & 1.42& 2.55&1.05 & 15.23 &19.05 & 12.36 &32.11&35.88&24.95 \\
ComplEx~\cite{ComplEx:16}&4.22 &5.97 &3.44 &19.10 &23.08 &12.99 & 32.91 &34.62 &24.67     \\
ConvE~\cite{conve:18}& \textbf{20.6} &\textbf{26.9}& 11.96 & 24.39&30.59& 18.51&33.02&39.78&24.45 \\
RotatE~\cite{sun2019rotate} &9.25 &11.83 &8.04 &25.96 &31.63 &23.34 & \textbf{58.32} & \textbf{60.66} & \textbf{51.85} \\
\hline
RC-Net~\cite{rcnet}      & 13.48  & 15.37  & \textbf{13.26} & 14.87 & 16.54& 14.63&14.69&16.34&14.41 \\ 
TransE+Line      & 12.17 &15.16 & 4.88 & 21.7 & 25.75& 8.81&26.76&31.65&10.97 \\ 
JointNRE~\cite{OpenNRE:18}&16.93 &20.74&11.39 &26.96 &31.54&21.24 &42.02&47.33&32.68 \\ 
\hline
MINERVA~\cite{GFAW:17} & 11.64 & 14.16 &8.93 & 25.16 &31.54  & 22.24& 43.80 & 44.70 & 34.62 \\
Two-Step &12.14 &16.5 &9.27 &21.66&31.50 & 19.82 &39.22&44.64&34.18\\
CPL (our method) & 15.19 & 18.00 & 10.87 & \textbf{26.81}  & \textbf{31.7}  & \textbf{23.80} & 43.25 & 49.50 & 33.52  \\ \hline
\end{tabular}
}
\vspace{-0.1cm}
\caption{\textbf{Performance comparison on the KG reasoning on the FB60K-NYT10 dataset.} We can observe similar performance trends as those on the UMLS-PubMed dataset.}
\label{overall-fb}
\vspace{-0.2cm}
\end{table*}

\subsection{Performance Comparison}
Performances of the KG reasoning of all the algorithms are given in Table \ref{overall-umls}, \ref{overall-fb} and Figure \ref{fig:fs}. We can draw conclusions as follows:

\smallskip
\noindent
\textbf{1. Triple ranking vs. path inference.}
CPL and MINERVA perform worse than triple-ranking methods when the size of KGs is small, while outperforms them significantly when adding more triples to the KGs (Figure \ref{fig:fs}). This is because the general and evidential paths for reasoning on sparse KGs are not enough, and path-based models cannot capture the underlying patterns.

\smallskip
\noindent
\textbf{2. CPL vs. joint embedding methods.} 
CPL is inferior to RC-net, TransE+Line, and JointNRE on small KG partitions because they are not path-based models and the connections on small KGs are too sparse. CPL outperforms them significantly on larger datasets. The reasons are two-fold : 1) the graphs are denser to provide enough reasoning paths for training; 2) other algorithms do not filter noisy text information in joint-training. 


\smallskip
\noindent
\textbf{3. CPL vs. other graph reasoning methods.} CPL outperforms MINERVA significantly because CPL makes use of relevant text information for prediction. MINERVA is better than CPL on full FB60K-NYT10 because the alignment between FB60K and NYT10 is very limited (Sec. \ref{data-alg}). The graph is dense at 100\%, and the benefits from the corpus information are indiscernible. 


\subsection{Performance Analysis}
\begin{figure}[tb]
\vspace{-0.1cm}
\includegraphics[width=210pt]{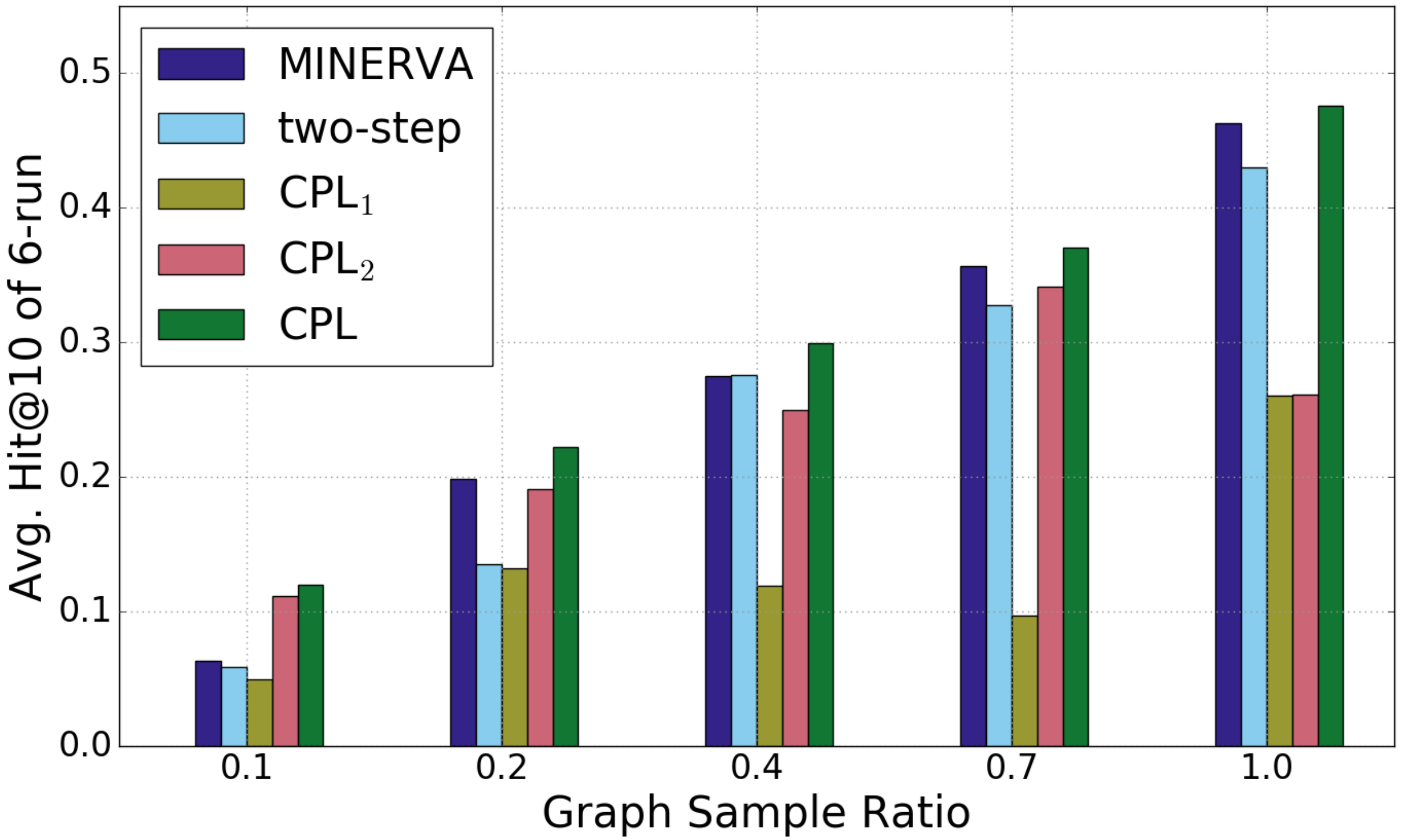}
\vspace{-0.2cm}
\caption{\textbf{Ablation Study on UMLS-PubMed dataset.} CPL$_1$ denotes CPL without adaptive sampling, and the extractor is frozen during training. CPL$_2$ denotes CPL without adaptive sampling. CPL denotes our proposed final model (with all the components).}
\label{fig:ablation}
\vspace{-0.2cm}
\end{figure}

\noindent
\textbf{1. Ablation Study on Model Components.}
In CPL, we apply multiple learning techniques to improve the performance, including collaboration, {replay memory}, and {adaptive sampling} as introduced in Sec.~\ref{implementation}. To show the effects of different components, we remove them one by one and train the respective model variants. In addition, to shown the effect of collaboration, we train a model variation with the parameters of the extractor frozen. The result is shown in Fig.~\ref{fig:ablation}.

From the result, we find that 1) replay memory is only effective when adaptive sampling is also enabled. This is because adaptive sampling solves the sparse positive sample problem to some extent. There are enough positive experiences for replay. 2) Collaboration improves performance significantly. CPL with a trainable extractor performs better than with a frozen extractor, which means the suggestions of the extractor can be improved by the reasoner's feedbacks. 3) The improvement of CPL over MINERVA reduces as we increase the KG size. This is because with more data for training, the graph becomes denser, and hence the contribution from texts will be diluted.

\smallskip
\noindent
\textbf{2. Effectiveness of Fact Selection.}
As mentioned above, Two-Step is the naive solution to OKGR. The best performing Two-Step model adds tens times more edges into the KG than CPL, whereas the Two-Step model's performance is inferior to CPL and MINERVA on all the datasets (Table \ref{overall-umls}, \ref{overall-fb}). The reasons are 1) most of the extracted edges used in the Two-Step model are noisy; 2) adding so many edges significantly enlarges the exploration space for reasoning.

\smallskip
\noindent
\textbf{2.} We perform a case study on the FB60K-NYT10 dataset to show the effectiveness of dynamically fact-filtering. We check the reasoning performance of the MINERVA and CPL periodically during the training. The results show that the extractor's contribution increases along with the training progress and the adaptive sampling can generate sufficient positive training experiences at the very beginning.

\begin{figure}[tb!]
\vspace{-0.2cm}
\includegraphics[width=216pt]{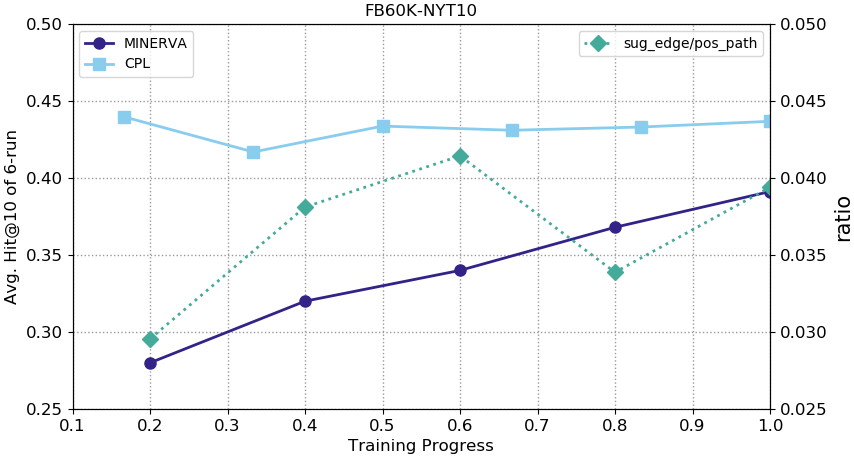}
\vspace{-0.1cm}
\caption{\textbf{KG reasoning performance change w.r.t. time.} sug\_edge/pos\_path means the ratio of positive edges suggested by the extractor w.r.t. the positive paths found by the reasoner.}
\label{fig:case1}
\vspace{-0.4cm}
\end{figure}

The result is shown in Figure \ref{fig:case1}. We find a few interesting points as follows: 
\textbf{1)} the sug\_edge/pos\_path ratio curve in Figure \ref{fig:case1} suggests that the extractor's contribution increases along with the training progress; 
\textbf{2)} CPL has a high initial performance because the adaptive sampling generates sufficient positive training experiences quickly. 
\textbf{3)} The valley shape in the performance curve is because the agent has not learned a stable exploring policy when the adaptive sampling stops, and the adaptive sampling somehow twisted the true pattern distribution in the dataset. But with a good start, the agent can explore on its own to approach the true distribution.

\subsection{Case Study of Reasoning Paths}
We randomly sample some reasoning paths from the inference results of CPL as examples.
Due to the space limit, please refer to the supplemental materials for these examples. 
These examples show 1) how the reasoner finds the path patterns for the respective relations; 2) how the reasoner finds the inference paths according to the patterns; 3) how the extractor suggests relevant edges for each positive paths; 4) how the extractor extracts the relevant facts from related sentences. In summary, these cases show how CPL performs interpretable knowledge graph reasoning (infer the query entity through semantics-related path searching) and how CPL performs interpretable fact-filtering (suggest edges w.r.t the learned reasoning path patterns).

%% file: 5-related.tex

\section{Related Work}
\vspace{-0.1cm}
\noindent
\textbf{Knowledge Graph Reasoning.}
Diverse approaches of embedding-based KG reasoning are presented, including linear models~\cite{TransE:11}, latent factor models~\cite{ComplEx:16}, matrix factorization models~\cite{DistMult:15} and convolutional neural networks~\cite{conve:18}. Performances of these methods are promising, but their predictions are barely interpretable. RL plays a crucial role in interpretable KG reasoning. MINERVA \cite{GFAW:17} and DeepPath \cite{DeepPath:17} employ policy networks; \cite{DeepPath:17} uses extra rule-based supervision. Multi-hop \cite{multi-lin2018} improves MINERVA via reward shaping and action drop-out.

\begin{figure}[tb]
\includegraphics[width=216pt]{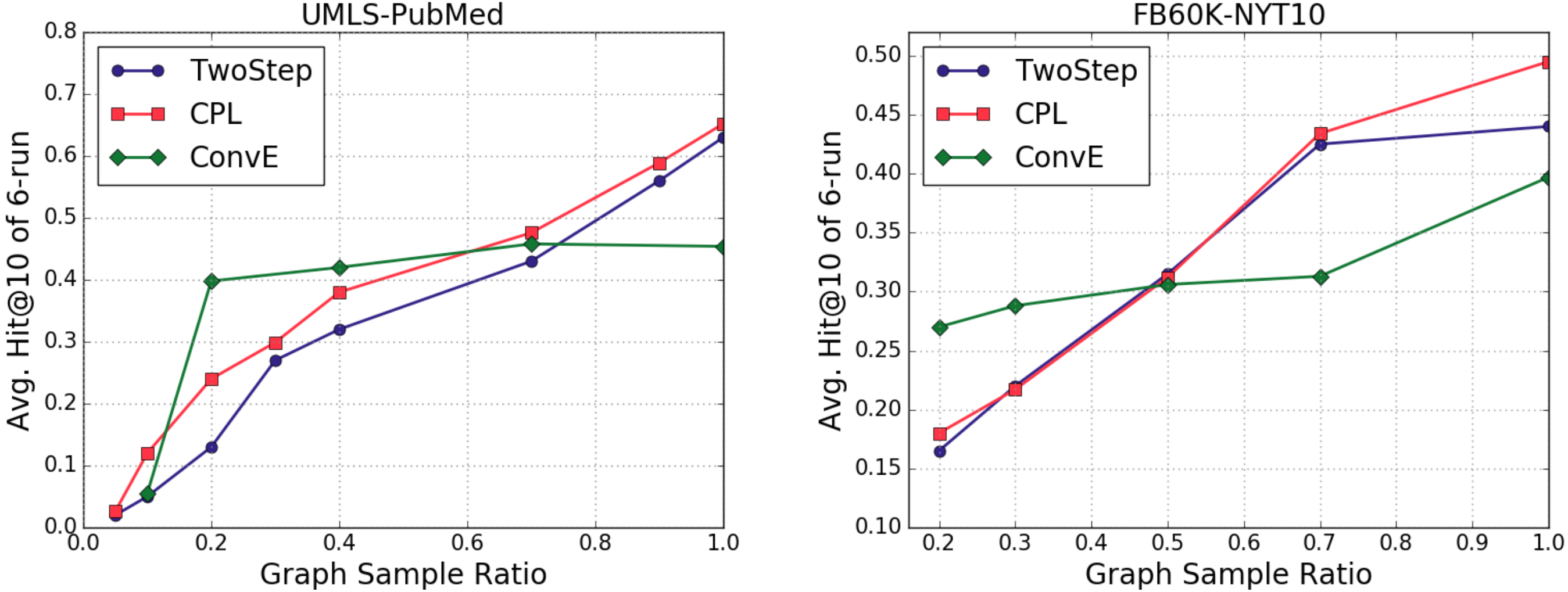}
\caption{\textbf{KG reasoning performance change w.r.t. the size of the graph.} Triple-ranking based methods perform pretty well on smaller partitions, but are soon surpassed by path-based KG reasoning methods with the increase of graph size.}
\label{fig:fs}
\vspace{-0.2cm}
\end{figure}

\smallskip
\noindent
\textbf{Joint Embedding of Text and KG.}
Joint embedding methods aim to unite text corpus and KG. Contrary to our focus, they mainly utilize KGs for better performances of other tasks. \cite{Danqi:15} focuses on fact-extraction on the corpus labeled via dependency parsing with the aids of KG and word embeddings, while \cite{THU:16} conducts the same task with the raw corpus text. As a newer joint model developed from \cite{THU:16}, \cite{OpenNRE:18} deals with fact extraction by employing the mutual attention. 

\smallskip
\noindent
\textbf{Open-World KG Completion.}
There are works focusing on similar topics as ours. \cite{shi2018open} defines an Open World KG Completion problem, in which they complete the KG with unseen entities.  \cite{friedman2019constrained} introduces the Open-World Probabilistic Databases, an analogy to KGs. Unlike our setting, they try to complete the KG with logical inferences without extra information. \cite{sun2018open} proposes an open, incomplete KB environment (or KG) with text corpora, but they focus on extracting answers from question-specific subgraphs.

%% file: 6-con.tex

\section{Conclusion}
In this paper, we focus on a new task named as Open Knowledge Graph Reasoning, which aims at boosting the knowledge graph reasoning with new knowledge extracted from the background corpus. We propose a novel and general framework, namely Collaborative Policy Learning, for this task. CPL trains two collaborative agents, the reasoner and fact extractor, which learns the path-reasoning policy and relevant-fact-extraction policy respectively. CPL can perform efficient interpretable reasoning on the KG and filtering of noisy facts. Experiments on two large real-world datasets demonstrate the strengths of CPL. Our work can cope with different path-finding modules such as MultHop with reward shaping by ConvE or RotatE and thus can improve its performance as the module improves.

\section{Acknowledgement}
This work has been supported in part by National Science Foundation SMA 18-29268, DARPA MCS and GAILA, IARPA BETTER, Schmidt Family Foundation, Amazon Faculty Award, Google Research Award, Snapchat Gift, JP Morgan AI Research Award, and China Scholarship Council. We would like to thank all the collaborators for their constructive feedbacks.

%% file: Supplemental.tex
\section{Supplemental Material}
\label{sec:supplemental}

\subsection{An Implementation of CPL}
\label{implement}
CPL is a general framework whose components, two agents, are all replaceable. In our experiments, we modify MINERVA \cite{GFAW:17} to construct the reasoner, and modify PCNN-ATT \cite{PCNN:16} to construct the fact extractor. Here we briefly introduce a specific implementation of CPL based on PCNN-ATT and MINERVA (see also Fig.~\ref{fig:JointModel} for illustration) in details.

\smallskip
\noindent
\textbf{Fact Extractor.}
PCNN-ATT is an effective Relation Extraction approach containing mainly two parts: the sentence encoder and the attention-selector. 
The sentence encoder encodes each sentence into a vector given the labeled entity pair and their positions in the sentence. We organize the sentences into sentence bags. The sentences in the same bag share the same entity-pair label ($x_\omega = (e^s,e^o)$). For each sentence bag, we modify PCNN-ATT to produces a predictive probability distribution over all relations in the vocabulary: $\Phi(\omega) = F_{pcnn}(x_\omega)$. 
Suppose at time step $t$ during the inference, the reasoner is at entity $e^t$. We need to suggest several edges pointing to different entities from $e^t$ to enrich the reasoner's action space. We use PCNN-ATT to make predictions on several sentence bags whose labels all contain $e^t$ and get a distribution set: $[\Phi(\omega_1), ... , \Phi(\omega_k)] = F_{pcnn}([x_{\omega_1}, ... , x_{\omega_K}]), w_k=(e^t, r^t_k, e^t_k), k\in[1,K]$. The distribution set can be seen as a score set over different edges. We can define a stochastic policy based on the scores by sampling the edges according to the scores. In the original PCNN-ATT setting, the score indicates the confidence of linking $e^s$ and $e_o$ w.r.t. the respective predicted relation. According to the previous reward definition, we can construct the policy distributions over all candidate edges based on these output scores via softmax and provide the extractor with the most relevant edges. 


\smallskip
\noindent
\textbf{Graph Reasoner.}
To extend MINERVA as our graph reasoner, we adopt random action drop-out (random dropping KG-edges) to unite the KG-edges and corpus-extracted edges into a joint action space of fixed size. Specifically, for a query triple $(e_s, r_q, e_q)$, it predicts $e_q$ through finding a path from $e_s$ to $e_q$ w.r.t. $r_q$. At time step $t$, the observed state $s^t$ is $(e^s,r_q,e^t,h^t)$ as defined before. The history information before $t$ is a sequence of edges, which is encoded into a vector with LSTM : $h^t=LSTM(h_{t-1},[r^{t}; e^{t}])$. The reasoner should select one edge from the joint action space defined above w.r.t. $e^t$. The MINERVA model $F_{mine}$ takes in the state embedding and output the softmax scores w.r.t. each action (out-edge). Then we adopt adaptive sampling (discussed below)  to select the action to proceed. 

\smallskip
\noindent
\textbf{Training pipeline.} A formal training algorithm is given in Algorithm \ref{train}.

\begin{algorithm}[t!h]
	\caption{CPL($G$, $C$, $b_r$, $b_e$, $p_l$, $e_a$, $e_m$)}
	\label{train}
	\begin{algorithmic}[1]
		\Require Knowledge graph $G$, corpus $C$, \# of batches training the reasoner $b_r$, \# of batches training the extractor $b_e$, hyper parameters for learning $p_l$, \# of epochs applying adaptive sampling $e_a$, maximal epochs $e_m$.
		\Ensure CPL model
		\State Initialize the reasoner and extractor.
		\State Register Adam optimizer with $p_l$.
		\For{$e$ = 0: $e_m$}
		\For{0 to max-batches}
		\If{$e<e_a$}
		\State Generate training sequences with \State adaptive sampling.
		\Else Generate training sequences with \State normal sampling.
		\EndIf
		\State Store the training sequences into \State replay memories. 
		\State Sample from the replay memory to \State train the reasoner for $b_r$ batches.
		\State Sample from the replay memory to \State train the reasoner for $b_r$ batches.
		\EndFor
		\EndFor
	\end{algorithmic}
\end{algorithm}
\subsection{Training techniques}
we will introduce a few techniques we use to increase training efficiency.

The lack of positive training samples is a common challenge for most RL algorithms. We use two techniques to accumulate positive experiences for the agents, \textit{model pre-training} and \textit{adaptive sampling}.

i) Model Pre-training. To get proper initialization, we pre-train the fact extractor and reasoner. In this way, at the beginning of the joint training, we can expect the agents to generate plausible experiences immediately.

ii) Adaptive Sampling. The policy learned by the agents can be regarded as a distribution of choosing certain actions given the states. Usually, we sample the actions multiple times according to the distribution to generate multiple experiences. In the pre-training stage, the reasoner is unaware of the facts in the texts. It tends to ignore the new facts suggested by the extractor. To facilitate interactions between two agents and encourage exploration, we reconstruct the distribution to ensure the extracted edges to be chosen with higher probability. Specifically, at time step $t$, the action space of the reasoner is the union of KG-edges and extracted edges, i.e., $A^t = \{(r,e)|(e^t,r,e)\in KG\} \cup$ $\{(r',e')|(e^t,r',e')\in$ corpus $C\}$. The reasoner will score all the actions in $A^t$, and we increase the scores of extracted edges adaptively so that they have higher priority over the KG-edges. Whereas we cannot keep this priority all the time, it twists the true data or pattern distribution. Hence after a number of iterations, we stop the adaptive sampling and use the immediate policy distribution for sampling.

To increase exploration efficiency, the fact extractor samples multiple edges given its learned policy to add to the reasoner's joint action space (Fig. \ref{fig:JointModel}). We collect the experiences with above techniques and store them into two replay memories \cite{replaymem} for two agents separately.

\subsection{Experiment Details}
\subsubsection{Datasets and Codes\footnote{The two datasets aforementioned in this paper and data pre-processing codes are in the supplementary materials and also available at 
\url{https://drive.google.com/file/d/1hCyPBjywpMuShRJCPKRjc7n2vHpxfetg/view?usp=sharing}. The codes in the supplemental material is our implementation of the CPL.}}
We study the datasets and find that the relation distributions of the two datasets are very imbalanced. There are not enough reasoning paths for some relation types. Moreover, some relations are meaningless and of no reasoning value. We select a subset of the relations for each dataset as the reasoning tasks. There are enough reasoning paths for the path-based models to learn on these relations. They are also pretty informative and widely concerned according to the opinions of the domain experts we interviewed. The details are in Table \ref{tab:relations}. Specifically, we first divide the dataset into train, validation, and test sets in the proportion $8:1:1$ randomly. Then we only keep the triples of the concerned relations in the validation and test set.

\subsubsection{Training Setup}
We list the parameter and experimental set-ups for all the algorithms in this section. The parameters not mentioned below have minor influences on the performance, so we follow the default configurations in their codes.

\textbf{ComplEx, DistMult, \&TransE}
We use the implementation from OpenKE \footnote{https://github.com/thunlp/OpenKE}. We set the embedding dimension as 100. We train each model for 600 iterations and 800 samples within each iteration. 

\textbf{RotatE}
We use the implementation supplied by its author \footnote{https://github.com/DeepGraphLearning/KnowledgeGraphEmbedding}. We set the embedding dimension as 100 (although the recommended value is 1000; we set this to avoid biases and training hurdles). We train each model for a total of 150k steps, with batch size of 256.  

\textbf{ConvE}
We use the public code for evaluation\footnote{https://github.com/TimDettmers/ConvE}. We set the embedding dimension as 200, training for 50 epochs. We use the same negative sampling ratio (i.e., 1:1) as what we use in the above OpenKE models for FB60K; and 1:all negative sampling for UMLS.

\textbf{Rc-net}
We use the code provided by the authors of paper \cite{rcnet}. We use all the default parameters except that we set the sample number as 48. In this way, we ensure that the training sample quantity used in Rc-net is the same as others.

\textbf{JointNRE}
We get the code from the authors of paper \cite{OpenNRE:18}. We use the PCNN-ATT as the sentence encoder and transE as the KG embedding method, which is the best-performing combination according to the authors. PCNN is trained for 20 iterations, during which the KG is trained by selecting 100 samples each batch, reaching 7,500 iterations at the end of training. Since the pre-trained word vector is a 50-dimension set, the embedding dimension is also 50. 

\textbf{LINE+TransE}
To train the word network and entity dictionary, the window is set to 5.  For the embedding part, embedding dimension is set to 50; 100 samples are selected in each epoch, while the number of epochs is stable at 1,000,000. 

\textbf{MINERVA}
We use the code \footnote{https://github.com/shehzaadzd/MINERVA} for evaluation. Since our Joint model approach requires MINERVA as a base model, we use the same embedding sizes and hidden sizes on the MINERVA training and our model training. To get our result we trained it for 400 iterations at a batch size of 64 samples on FB60K. We set the iteration-number to 1000 and batch-size to 64 for UMLS.

To better reflect the models' capabilities, all models related to MINERVA are added reverse edge triples. Considering the inevitable fluctuations of this reinforcement learning model, we use three random keys 55, 83 and 5583 to initiate training and reach an average result for the three runs.

\textbf{Our model}
In total we train 400 iterations for FB60K (considering time factors) and 1000 for UMLS; For first 200 iterations, we use BFS to search positive paths with higher priority on PCNN-ATT suggested edges. In each BFS iteration, 100 samples are selected. The learning rate is set to 0.001, and the batch size is 64, the same as MINERVA.

\subsection{Case Study}
\smallskip
\noindent
\textbf{1.} Two-step is the naive solution to OKGR. For the two-step model, we filter the corpus-edges with the output scores (in [0,1]) of PCNN-ATT. 0 means adding all the edges to the KG, while 1 means adding nothing. We find the best threshold (producing the best reasoning model) for UMLS-PubMed is 0.5 and 0 for FB60K-NYT10. Two-step adds about 85,000 edges to UMLS and 90,000 to FB60K under the corresponding thresholds, whereas CPL adds about 8,000 edges to UMLS and 1,500 for FB60K. 

The two-step model performance is inferior to CPL and MINERVA on all the datasets (Table \ref{overall-umls}, \ref{overall-fb}). The reasons are that 1) most of the extracted edges use in the two-step model are noises; 2) adding so many edges significantly enlarges the explore space for reasoning. Selecting the correct out-edge at each step becomes more difficult. Lack sufficient positive experiences, with same iterations, the two-step model cannot learn the underlying patterns well.

\begin{figure*}[]
	\centering
	\subfigure{\includegraphics[height=700pt, width=400pt]{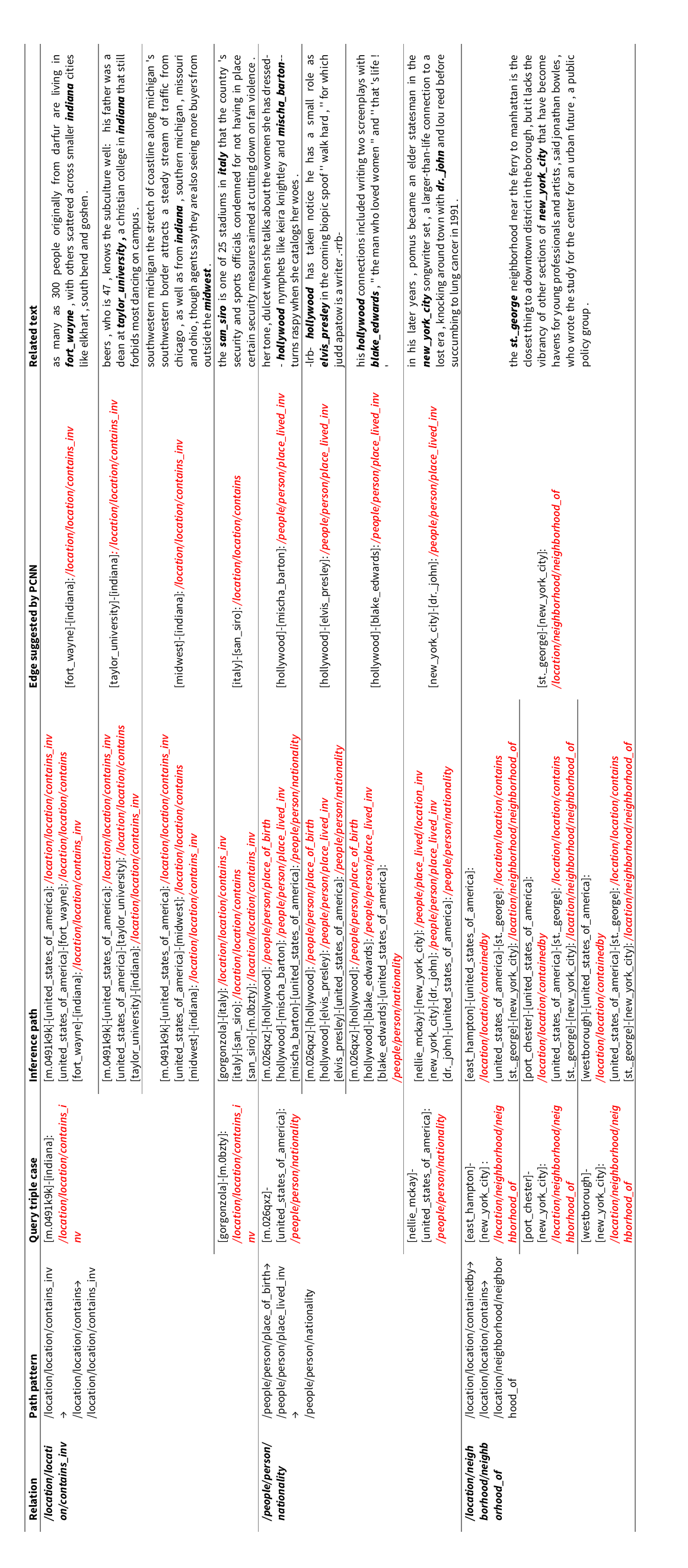}}
\caption{\textbf{A Case study on discovered paths on FB60K-NYT10.}We randomly pick three relations and show how CPL performs reasoning based on the KG and text corpus. Red texts are the relations. $[xxx]$-$[xxx]$ represents [subject entity]-[object entity]. The bold italic words in the sentences means where we extract the relations.}
\label{fig:casestudy}
\end{figure*}

\smallskip
\noindent
\textbf{3.} Figure \ref{fig:casestudy} shows the inference cases randomly sampled from the FB60K-NYT10 dataset. We select three relations and randomly sample several query cases from the test set. We track down the inference paths for each query case and mark the edges suggested by the extractor. Further, we track back to the raw text data to pick out the sentences from which the extractor extract the relevant facts. For example, for the query triple (gorgonzola, /location/location/contains\_inv, m.0bzty), the concerned relation is ``/location/location/contains\_inv''. A possible pattern to infer the relation is ``/location/location/contains\_inv'' $\land$ ``/location/location/contains'' $\land$ 
``/location/location/contains\_inv'' $\rightarrow$ ``/location/location/contains\_inv''. The specific path found by the reasoner is (gorgonzola, /location/location/contains\_inv, Italy) $\rightarrow$ (Italy, /location/location/contains, san\_siro) $\rightarrow$ (san\_siro, /location/location/contains\_inv, m.0bzty). Among them, edge (Italy, /location/location/contains, san\_siro) is a new edge suggested by the extractor, which is extracted from the sentence ``the san\_siro is one of 25 stadiums in italy that the country ’s security and sports officials condemned for not having in place certain security measures aimed at cutting down on fan violence ''.

\begin{table*}[tb]
\centering
\scalebox{0.68}{
\begin{tabular}{lcccccccccccc}
\hline
Model / Dataset     & \multicolumn{3}{c}{\textbf{20\%}} & \multicolumn{3}{c}{\textbf{50\%}} & \multicolumn{3}{c}{\textbf{100\%}} \\ 
   & $\sigma(Hits@5)$ & $\sigma(Hits@10)$ & $\sigma(MRR)$ & $\sigma(Hits@5)$ & $\sigma(Hits@10)$ & $\sigma(MRR)$ & $\sigma(Hits@5)$ & $\sigma(Hits@10)$ & $\sigma(MRR)$        \\ 
\hline\hline
TransE~\cite{TransE:11}     &0.0013&0.0011&0.0017&0.0015&0.0014&0.0014&0.0015&0.0015&0.0011 \\
DisMult~\cite{DistMult:15}  &0.0015&0.0015&0.0017&0.0010&0.0015&0.0017&0.0017&0.0020&0.0012 \\
ComplEx~\cite{ComplEx:16}   &0.0040&0.0030&0.0030&0.0013&0.0025&0.0029&0.0024&0.0026&0.0027 \\
ConvE~\cite{conve:18}       &0.0031&0.0043&0.0026&0.0020&0.0027&0.0019&0.0032&0.0026&0.0027\\
\hline
RC-Net~\cite{rcnet}         &0.0009&0.0016&0.0016&0.0013&0.0016&0.0015&0.0018&0.0007&0.0017\\ 
TransE+Line                 &0.0015&0.0013&0.0008&0.0014&0.0005&0.0013&0.0013&0.0012&0.0014\\ 
JointNRE~\cite{OpenNRE:18}  &0.0015&0.0012&0.0016&0.0015&0.0016&0.0007&0.0016&0.0012&0.0015\\ 
\hline
MINERVA~\cite{GFAW:17}      &0.0100&0.0118&0.0148&0.0856&0.1009&0.0550&0.0974&0.0849&0.1253\\
Two-Step                    &0.0140&0.0137&0.0095&0.0290&0.0279&0.0309&0.0343&0.0368&0.0614\\
CPL (our method)            &0.0028&0.0017&0.0044&0.0131&0.0033&0.0547&0.0040&0.0010&0.0227\\ \hline
\end{tabular}
}
\vspace{-0.1cm}
\caption{\textbf{Performance variance of KG reasoning on the FB60K-NYT10 dataset.} Reinforcement learning methods do suffer from variances between different runs.}
\label{sigma-fb}
\vspace{-0.2cm}
\end{table*}

\begin{table*}[ht]
\centering
\scalebox{0.68}{
\begin{tabular}{lcccccccc}
\hline
Model / Dataset    & \multicolumn{2}{c}{\textbf{20\%}} & \multicolumn{2}{c}{\textbf{40\%} } & \multicolumn{2}{c}{\textbf{70\%}} & \multicolumn{2}{c}{\textbf{100\%}} \\ 
          & $\sigma(Hits@5)$ & $\sigma(Hits@10)$ & $\sigma(Hits@5)$ & $\sigma(Hits@10)$ & $\sigma(Hits@5)$ & $\sigma(Hits@10)$ & $\sigma(Hits@5)$ & $\sigma(Hits@10)$     \\ \hline\hline
TransE~\cite{TransE:11}     & 0.0027&0.0023&0.0035&0.0039&0.0034&0.0032&0.0029&0.0020\\
DisMult~\cite{DistMult:15}  & 0.0020&0.0034&0.0031&0.0017&0.0029&0.0032&0.0033&0.0022\\
ComplEx~\cite{ComplEx:16}   & 0.0026&0.0022&0.0035&0.0029&0.0036&0.0007&0.0016&0.0024\\
ConvE~\cite{conve:18}       & 0.0028&0.0027&0.0030&0.0028&0.0025&0.0033&0.0029&0.0020\\
\hline
RC-Net~\cite{rcnet}         & 0.0024&0.0030&0.0013&0.0030&0.0030&0.0014&0.0026&0.0022\\ 
TransE+Line                 & 0.0027&0.0040&0.0026&0.0029&0.0024&0.0013&0.0036&0.0013\\ 
JointNRE~\cite{OpenNRE:18}  & 0.0008&0.0016&0.0026&0.0028&0.0028&0.0034&0.0028&0.0019\\ 
\hline
MINERVA~\cite{GFAW:17}      & 0.0171&0.0195&0.0327&0.0217&0.0565&0.0499&0.0575&0.0678\\
Two-Step                    & 0.0072&0.0088&0.0193&0.0178&0.0217&0.0025&0.0021&0.0094\\
CPL (our method)            & 0.0155&0.0020&0.0090&0.0031&0.0166&0.0028&0.0155&0.0033\\ \hline
\end{tabular}
}
\caption{\textbf{Performance variance of KG reasoning on the UMLS-PubMed dataset.}}
\label{sigma-umls}
\end{table*}

%% file: acl2019.bbl
\begin{thebibliography}{21}
\expandafter\ifx\csname natexlab\endcsname\relax\def\natexlab#1{#1}\fi

\bibitem[{Bordes et~al.(2011)Bordes, Weston, Collobert, Bengio
  et~al.}]{TransE:11}
Antoine Bordes, Jason Weston, Ronan Collobert, Yoshua Bengio, et~al. 2011.
\newblock Learning structured embeddings of knowledge bases.
\newblock In \emph{AAAI}, volume~6, page~6.

\bibitem[{Das et~al.(2017)Das, Dhuliawala, Zaheer, Vilnis, Durugkar,
  Krishnamurthy, Smola, and McCallum}]{GFAW:17}
Rajarshi Das, Shehzaad Dhuliawala, Manzil Zaheer, Luke Vilnis, Ishan Durugkar,
  Akshay Krishnamurthy, Alex Smola, and Andrew McCallum. 2017.
\newblock Go for a walk and arrive at the answer: Reasoning over paths in
  knowledge bases using reinforcement learning.
\newblock \emph{arXiv preprint arXiv:1711.05851}.

\bibitem[{Dettmers et~al.(2018)Dettmers, Minervini, Stenetorp, and
  Riedel}]{conve:18}
Tim Dettmers, Pasquale Minervini, Pontus Stenetorp, and Sebastian Riedel. 2018.
\newblock Convolutional 2d knowledge graph embeddings.
\newblock In \emph{Thirty-Second AAAI Conference on Artificial Intelligence}.

\bibitem[{Friedman and Broeck(2019)}]{friedman2019constrained}
Tal Friedman and Guy Van~den Broeck. 2019.
\newblock On constrained open-world probabilistic databases.
\newblock \emph{arXiv preprint arXiv:1902.10677}.

\bibitem[{Han et~al.(2016)Han, Liu, and Sun}]{THU:16}
Xu~Han, Zhiyuan Liu, and Maosong Sun. 2016.
\newblock Joint representation learning of text and knowledge for knowledge
  graph completion.
\newblock \emph{arXiv preprint arXiv:1611.04125}.

\bibitem[{Han et~al.(2018)Han, Liu, and Sun}]{OpenNRE:18}
Xu~Han, Zhiyuan Liu, and Maosong Sun. 2018.
\newblock Neural knowledge acquisition via mutual attention between knowledge
  graph and text.
\newblock In \emph{Proceedings of AAAI}.

\bibitem[{Lin et~al.(2018)Lin, Socher, and Xiong}]{multi-lin2018}
Xi~Victoria Lin, Richard Socher, and Caiming Xiong. 2018.
\newblock Multi-hop knowledge graph reasoning with reward shaping.
\newblock \emph{arXiv preprint arXiv:1808.10568}.

\bibitem[{Lin et~al.(2016)Lin, Shen, Liu, Luan, and Sun}]{PCNN:16}
Yankai Lin, Shiqi Shen, Zhiyuan Liu, Huanbo Luan, and Maosong Sun. 2016.
\newblock Neural relation extraction with selective attention over instances.
\newblock In \emph{Proceedings of the 54th Annual Meeting of the Association
  for Computational Linguistics (Volume 1: Long Papers)}, volume~1, pages
  2124--2133.

\bibitem[{Mintz et~al.(2009)Mintz, Bills, Snow, and Jurafsky}]{Mintz:09}
Mike Mintz, Steven Bills, Rion Snow, and Dan Jurafsky. 2009.
\newblock \href {http://dl.acm.org/citation.cfm?id=1690219.1690287} {Distant
  supervision for relation extraction without labeled data}.
\newblock In \emph{Proceedings of the Joint Conference of the 47th Annual
  Meeting of the ACL and the 4th International Joint Conference on Natural
  Language Processing of the AFNLP: Volume 2 - Volume 2}, ACL '09, pages
  1003--1011, Stroudsburg, PA, USA. Association for Computational Linguistics.

\bibitem[{Mnih et~al.(2013)Mnih, Kavukcuoglu, Silver, Graves, Antonoglou,
  Wierstra, and Riedmiller}]{replaymem}
Volodymyr Mnih, Koray Kavukcuoglu, David Silver, Alex Graves, Ioannis
  Antonoglou, Daan Wierstra, and Martin Riedmiller. 2013.
\newblock Playing atari with deep reinforcement learning.
\newblock \emph{arXiv preprint arXiv:1312.5602}.

\bibitem[{Shi and Weninger(2018)}]{shi2018open}
Baoxu Shi and Tim Weninger. 2018.
\newblock Open-world knowledge graph completion.
\newblock In \emph{Thirty-Second AAAI Conference on Artificial Intelligence}.

\bibitem[{Socher et~al.(2013)Socher, Chen, Manning, and
  Ng}]{socher2013reasoning}
Richard Socher, Danqi Chen, Christopher~D Manning, and Andrew Ng. 2013.
\newblock Reasoning with neural tensor networks for knowledge base completion.
\newblock In \emph{Advances in neural information processing systems}, pages
  926--934.

\bibitem[{Sun et~al.(2018)Sun, Dhingra, Zaheer, Mazaitis, Salakhutdinov, and
  Cohen}]{sun2018open}
Haitian Sun, Bhuwan Dhingra, Manzil Zaheer, Kathryn Mazaitis, Ruslan
  Salakhutdinov, and William~W Cohen. 2018.
\newblock Open domain question answering using early fusion of knowledge bases
  and text.
\newblock \emph{arXiv preprint arXiv:1809.00782}.

\bibitem[{Sun et~al.(2019)Sun, Deng, Nie, and Tang}]{sun2019rotate}
Zhiqing Sun, Zhi-Hong Deng, Jian-Yun Nie, and Jian Tang. 2019.
\newblock Rotate: Knowledge graph embedding by relational rotation in complex
  space.
\newblock \emph{arXiv preprint arXiv:1902.10197}.

\bibitem[{Tang et~al.(2015)Tang, Qu, Wang, Zhang, Yan, and Mei}]{tang2015line}
Jian Tang, Meng Qu, Mingzhe Wang, Ming Zhang, Jun Yan, and Qiaozhu Mei. 2015.
\newblock Line: Large-scale information network embedding.
\newblock In \emph{Proceedings of the 24th international conference on world
  wide web}, pages 1067--1077. International World Wide Web Conferences
  Steering Committee.

\bibitem[{Toutanova et~al.(2015)Toutanova, Chen, Pantel, Poon, Choudhury, and
  Gamon}]{Danqi:15}
Kristina Toutanova, Danqi Chen, Patrick Pantel, Hoifung Poon, Pallavi
  Choudhury, and Michael Gamon. 2015.
\newblock Representing text for joint embedding of text and knowledge bases.
\newblock In \emph{Proceedings of the 2015 Conference on Empirical Methods in
  Natural Language Processing}, pages 1499--1509.

\bibitem[{Trouillon et~al.(2016)Trouillon, Welbl, Riedel, Gaussier, and
  Bouchard}]{ComplEx:16}
Th{\'e}o Trouillon, Johannes Welbl, Sebastian Riedel, {\'E}ric Gaussier, and
  Guillaume Bouchard. 2016.
\newblock Complex embeddings for simple link prediction.
\newblock In \emph{International Conference on Machine Learning}, pages
  2071--2080.

\bibitem[{Williams(1992)}]{Reinforce:1999}
Ronald~J Williams. 1992.
\newblock Simple statistical gradient-following algorithms for connectionist
  reinforcement learning.
\newblock \emph{Machine learning}, 8(3-4):229--256.

\bibitem[{Xiong et~al.(2017)Xiong, Hoang, and Wang}]{DeepPath:17}
Wenhan Xiong, Thien Hoang, and William~Yang Wang. 2017.
\newblock Deeppath: A reinforcement learning method for knowledge graph
  reasoning.
\newblock \emph{arXiv preprint arXiv:1707.06690}.

\bibitem[{Xu et~al.(2014)Xu, Bai, Bian, Gao, Wang, Liu, and Liu}]{rcnet}
Chang Xu, Yalong Bai, Jiang Bian, Bin Gao, Gang Wang, Xiaoguang Liu, and
  Tie-Yan Liu. 2014.
\newblock Rc-net: A general framework for incorporating knowledge into word
  representations.
\newblock In \emph{Proceedings of the 23rd ACM international conference on
  conference on information and knowledge management}, pages 1219--1228. ACM.

\bibitem[{Yang et~al.(2014)Yang, Yih, He, Gao, and Deng}]{DistMult:15}
Bishan Yang, Wen-tau Yih, Xiaodong He, Jianfeng Gao, and Li~Deng. 2014.
\newblock Embedding entities and relations for learning and inference in
  knowledge bases.
\newblock \emph{arXiv preprint arXiv:1412.6575}.

\end{thebibliography}
